\documentclass[11pt]{article}
\usepackage[preprint]{acl}
\usepackage{times}
\usepackage{latexsym}
\usepackage[T1]{fontenc}
\usepackage[utf8]{inputenc}
\usepackage{microtype}
\usepackage{amsmath}
\usepackage{amssymb}
\usepackage{booktabs}
\usepackage{graphicx}
\usepackage{tikz}
\usetikzlibrary{positioning,arrows.meta,fit,backgrounds,calc}
\usepackage{enumitem}
\usepackage{xcolor}
\usepackage{listings}
\usepackage{upquote}

\DeclareUnicodeCharacter{0394}{\ensuremath{\Delta}}

\newcommand{\sysname}{our framework}

\title{Patching the Blind Spots: Reversible LLM Rule Overlay\\
       for Frozen Fraud Detectors}

\author{%
     \textbf{Bo Qu\textsuperscript{\ensuremath{\dagger}}\thanks{Corresponding Author.}},
      \textbf{Zhurong Wang\textsuperscript{\ensuremath{\dagger}}},
    \textbf{Frank Zahradnik\textsuperscript{\ensuremath{\dagger}}},
     \textbf{Ming Ouyang\textsuperscript{\ensuremath{\dagger}}}
    \\
     \textsuperscript{\ensuremath{\dagger}}eBay,
     \\
\texttt{{boqu@ebay.com}}
  }

\begin{document}
\maketitle
\begin{abstract}
Gradient-boosted trees dominate industrial tabular fraud classification, but even a well-trained model
has blind spots, and retraining a deployed model is not always safe. We bring LLM reasoning to patching
such a model without modifying it: LLMs read its tree digest and SHAP attributions, reason about the
fraud it misses, and propose rules anchored on features it underweights, screened and calibrated
into a reversible log-lift overlay at the production cut-off. Single-LLM benefit is
high-variance and inverts between datasets, so we pool three diverse backbones by logit averaging,
trading peak gain for a bounded worst case. On an experimental detector trained on
sampled, selection-biased real-world e-commerce data, tuned continued training regresses, whereas the overlay gains on
average $+2.26\%$ recall at equal precision ($\Delta$R@P) and $+1.60\%$ precision at equal recall
($\Delta$P@R). Replayed offline on a legacy production detector, the overlay,
calibrated on one month and frozen, stays more precise at equal recall for seven months ($+5.64\%$ $\Delta$P@R). On seven bias-shift variants of
a public benchmark it raises $\Delta$P@R on all and keeps $\Delta$R@P near neutral. LLM
reasoning over a model's diagnostics can improve a deployed tabular classifier through readable
predicates naming the fraud they cover.
\end{abstract}

\section{Introduction}
\label{sec:intro}

Gradient-boosted trees dominate tabular fraud classification
\citep{ke2017lightgbm,chen2016xgboost}. Trained once on historical labelled transactions, they
degrade under the temporal distribution shift ubiquitous in financial data
\citep{quinonero2009shift,gama2014drift,dalpozzolo2018fraud}: new fraud patterns emerge, the
population shifts, and discrimination quietly erodes with no visible retraining signal.

In production, such a detector also has \emph{loopholes}: fraud patterns it scores too low, which
persist after deployment. Closing them is constrained in two ways in our setting. The model's scores
and cut-offs drive downstream decisions, so changing the model means a redeployment. Its training
data are selection-biased: a rule engine and a screening model filter transactions before they are
recorded, so training sees \emph{approved} traffic only, while the model serves the full population
(a small unbiased cohort is kept for evaluation; Section~\ref{sec:setup}). Patterns that rarely pass
the filters are under-represented, and the usual fix---continued training on a fresh window---fits
the same bias: in our experiments it regresses on the unbiased cohort (Section~\ref{sec:results}).
The practical question is how to find a frozen model's loopholes and improve how it generalizes,
without modifying it.

\paragraph{Key idea: LLMs reason about the frozen GBM's blind spots and patch them proactively.}
Instead of updating the model, we let large language models read its diagnostics (SHAP attributions
and a digest of its trees), reason about how fraud evades it, and propose human-readable rules for the
fraud it misses; the adaptation window only screens and calibrates them, so no model parameter changes.
Generation is steered \emph{through the model's weak dimensions}: each rule's primary anchor must be a
feature the model underweights by SHAP. The rules are calibrated into a band-conditioned log-lift
overlay on the \textbf{frozen} base---additive in logit space, hence reversible---and mapped back to
the production score scale, so the overlay runs at the existing cut-off. Every amendment is a
human-auditable predicate, in line with calls for interpretable high-stakes models
\citep{rudin2019interpretable}.

Prior work uses LLMs on tabular data either as the predictor itself, on serialized rows
\citep{hegselmann2023tabllm,dinh2022lift}, or to engineer features and rules for a model that is
then trained or retrained: CAAFE generates features from dataset descriptions
\citep{hollmann2023caafe}, FeatLLM turns LLM-written rules into features for few-shot learning
\citep{han2024featllm}, and OCTree refines feature-generation rules with decision-tree feedback
\citep{nam2024octree}. Closest to us, DeLTa has an LLM rewrite a tree ensemble's rules into one that
corrects the ensemble's outputs \citep{ye2025delta}. These methods target accuracy on public
benchmarks, where the predictor can be rebuilt freely. Our setting differs in three ways: the model is
deployed and frozen, so the rules act as a reversible overlay at its existing cut-off; generation is
anchored on the features the model underweights rather than on refining its existing tree rules; and the
training data are selection-biased, so the gain must hold on a population the model was not trained
on.

As the rule proposer, a single LLM produces high-variance outcomes not predictable from in-distribution
signal: the backbone that performs best on the real-world generalization test is among the worst on
a public benchmark, and vice versa (Section~\ref{sec:results}). We therefore prompt three
architecturally diverse backbones \citep{wei2022cot,deepseek2025r1,qwen2026qwen35,gemma2026gemma4} across
independent shots and average all retained shots in logit space \citep{dietterich2000ensemble}, with
no dataset-specific selection.

\subsection*{Contributions}

\begin{enumerate}[nosep,leftmargin=1.2em]
  \item \textbf{LLM reasoning to patch a frozen, deployed model.} LLMs propose rules from the
    detector's diagnostics, steered through its SHAP-underweighted dimensions and calibrated into a
    reversible overlay at the production cut-off, with every parameter fixed before deployment.
  \item \textbf{Evidence under selection bias, drift and designed shift.} On an unbiased cohort,
    continued training regresses while the overlay gains recall at equal precision; on a legacy
    production model, the overlay, calibrated on one month, stays more precise at equal recall for the
    next seven;
    on seven bias-shift variants it raises precision at equal recall on every variant. Pooling three LLMs bounds the high,
    dataset-dependent variance of single-LLM benefit without oracle selection
    (Section~\ref{sec:results}).
\end{enumerate}

\section{Method}
\label{sec:method}

\sysname\ adds a calibrated, LLM-generated rule overlay to a frozen gradient-boosted classifier.
Figure~\ref{fig:framework} summarizes the pipeline, from a model-derived prompt context through
LLM proposal, empirical screening, band-conditioned calibration, and a reversible overlay (top),
and how the overlay is deployed beside the production score (bottom).

\definecolor{colCtx}{RGB}{214,229,252}    %
\definecolor{colLLM}{RGB}{252,238,210}    %
\definecolor{colScreen}{RGB}{210,245,218} %
\definecolor{colCalib}{RGB}{237,226,250}  %
\definecolor{colOv}{RGB}{215,242,230}     %

\begin{figure*}[t]
\centering
\resizebox{\textwidth}{!}{%
\begin{tikzpicture}[
  >=Latex,
  node distance=0.45cm,
  box/.style={draw, rounded corners=3pt, align=center, inner sep=5pt,
              font=\footnotesize},
  mbox/.style={draw, rounded corners=2pt, align=center, inner sep=2pt,
               font=\scriptsize, fill=colLLM},
  arr/.style={->, thick},
  darr/.style={->, dashed, semithick, gray!65!black},
  anarr/.style={->, thin, gray!55!black, shorten >=2pt, shorten <=2pt},
]

\node[box, fill=colCtx, text width=1.95cm, minimum height=2.3cm] (ctx) {%
  \textbf{Prompt}\\
  \textbf{Context}\\[3pt]
  \scriptsize domain primer\\[2pt]
  model digest\\
  (splits, gain,\\
  paths)\\[2pt]
  SHAP gap $+$\\
  lift tables%
};
\node[font=\scriptsize, align=center, text width=2.05cm, text=gray!68!black,
      below=0.35cm of ctx] (ctxa)
  {\textit{summaries of the}\\
   \textit{frozen GBM;}\\
   \textit{model unchanged}};
\draw[anarr] (ctx.south) -- (ctxa.north);

\node[mbox, right=0.8cm of ctx, yshift=0.95cm, text width=1.9cm] (llm1) {%
  \textbf{LLM-CoT}\\
  chain-of-thought%
};
\node[mbox, below=0.22cm of llm1, text width=1.9cm] (llm2) {%
  \textbf{LLM-SlW}\\
  sliding-window%
};
\node[mbox, below=0.22cm of llm2, text width=1.9cm] (llm3) {%
  \textbf{LLM-MoE}\\
  sparse MoE%
};
\begin{pgfonlayer}{background}
  \node[draw, dashed, rounded corners=5pt, fill=colLLM!30,
        fit=(llm1)(llm2)(llm3), inner sep=7pt] (llmg) {};
\end{pgfonlayer}
\node[font={\scriptsize\bfseries}, anchor=south west]
  at ($(llmg.north west) + (3pt,2pt)$)
  {3 open-weight LLMs $\times$ 10 independent shots};

\node[box, fill=colScreen, right=0.85cm of llmg,
      text width=2.05cm, minimum height=2.3cm] (scr) {%
  \textbf{Support}\\
  \textbf{Screen}\\[3pt]
  \scriptsize compiles, fires,\\
  lift $>1$;\\[2pt]
  per band: $\geq 3$\\
  fires, lift $>1$%
};

\node[box, fill=colCalib, right=0.75cm of scr,
      text width=2.05cm, minimum height=2.3cm] (cal) {%
  \textbf{Band-Cond.}\\
  \textbf{Log-Lift}\\
  \textbf{Calibration}\\[3pt]
  \scriptsize log-lift residual\\
  per alert-rate\\
  band%
};

\node[box, fill=colOv, right=0.75cm of cal,
      text width=2.05cm, minimum height=2.3cm] (ov) {%
  \textbf{Logit-Space}\\
  \textbf{Averaging}\\
  $+$\,\textbf{Overlay}\\[3pt]
  \scriptsize uniform avg.\\
  across retained\\
  shots \& models;\\
  \emph{additive}\\
  overlay%
};
\node[font=\scriptsize, align=center, text width=2.2cm, text=gray!68!black,
      below=0.35cm of ov] (ova)
  {\textit{reversible: base GBM}\\
   \textit{never modified;}\\
   \textit{disable w/o redeploy}};
\draw[anarr] (ov.south) -- (ova.north);

\node[box, fill=white, right=0.75cm of ov,
      text width=1.55cm, minimum height=2.2cm] (out) {%
  \textbf{Final}\\
  \textbf{Score}\\[3pt]
  \scriptsize frozen GBM\\
  $+$ overlay%
};

\draw[arr] (ctx.east) -- (llmg.west);
\draw[arr] (llmg.east) --
  node[above, font=\scriptsize, fill=white, inner sep=1.5pt]{candidate rules}
  (scr.west);
\draw[arr] (scr.east) -- (cal.west);
\draw[arr] (cal.east) -- (ov.west);
\draw[arr] (ov.east) -- (out.west);

\end{tikzpicture}%
}
\par\medskip
\resizebox{\textwidth}{!}{%
\begin{tikzpicture}[
  >=Latex,
  box/.style={draw, rounded corners=3pt, align=center, inner sep=3.5pt, font=\footnotesize, minimum height=1.05cm},
  arr/.style={->, thick},
  oarr/.style={->, thick, teal!50!black},
]
\node[font=\scriptsize\bfseries, anchor=west, text=gray!60!black] at (-0.9, 1.55) {PRODUCTION (unchanged)};
\node[font=\scriptsize\bfseries, anchor=west, text=teal!50!black] at (-0.9, -1.75) {OVERLAY (same cut-off $t_b$)};
\node[box, fill=white, text width=1.65cm] (x) at (0,0) {\textbf{Transaction}\\$\mathbf{x}$};
\node[box, fill=colCtx, text width=2.6cm] (f) at (3.1,0.85) {\textbf{Frozen model} $f$\\\scriptsize production score $s(\mathbf{x})$};
\node[box, fill=white, text width=2.3cm] (pa) at (18.2,0.85) {\textbf{Production alert}\\\scriptsize $s(\mathbf{x})\ge t_b$};
\node[box, fill=colCalib, text width=2.6cm] (r) at (3.1,-0.85) {\textbf{Rule pools} $\mathcal{K}_{j,b}$\\\scriptsize $R_k(\mathbf{x})$, $\log\mathrm{lift}_k$};
\node[box, fill=colOv, text width=2.85cm] (u) at (6.9,-0.85) {\textbf{Base percentile}\\\scriptsize $p(\mathbf{x})=F_A\bigl(s(\mathbf{x})\bigr)$};
\node[box, fill=colOv, text width=3.9cm] (z) at (10.85,-0.85) {\textbf{Overlay score}\\\scriptsize $z_b(\mathbf{x})=\tfrac{1}{|S|}\sum_{j}\operatorname{logit}p^*_{j,b}(\mathbf{x})$};
\node[box, fill=colOv, text width=3.35cm] (st) at (14.85,-0.85) {\textbf{Scaled overlay score}\\\scriptsize $\tilde s_b=\max\{t: z_b\ge\tau_b(t)\}$};
\node[box, fill=colOv, text width=2.3cm] (oa) at (18.2,-0.85) {\textbf{Overlay alert}\\\scriptsize $\tilde s_b(\mathbf{x})\ge t_b$};
\draw[arr] (x.east) -- ++(0.25,0) |- (f.west);
\draw[arr] (x.east) -- ++(0.25,0) |- (r.west);
\draw[arr] (f.east) -- (pa.west);
\draw[oarr] ($(f.east)+(0.45,0)$) |- (u.west);
\draw[oarr] (u.east) -- (z.west);
\draw[oarr] (r.south) -- ++(0,-0.3) -| (z.south);
\draw[oarr] (z.east) -- (st.west);
\draw[oarr] (st.east) -- (oa.west);
\node[font=\scriptsize, text=gray!65!black, anchor=west] at (5.4,-2.05)
  {\textit{Fitted once on the adaptation window $A$, then frozen:} percentile table $F_A$; lifts; cut-off table $\tau_b(t)$, which alerts as many rows of $A$ as $s\ge t$.};
\node[font=\scriptsize, text=gray!65!black, anchor=west] at (5.4,-2.42)
  {\textit{Rules off} $\Rightarrow \tilde s_b = s$\textit{: the overlay is switched off without redeploying $f$.}};
\end{tikzpicture}}
\caption{\textbf{Top: framework overview (left to right).}
  A prompt context built from frozen base-model outputs drives three open-weight LLMs; emitted rules
  are support-screened and calibrated as band-conditioned log-lift
  residuals, giving an \emph{additive, reversible} overlay on the frozen base score.
  \textbf{Bottom: deployment.} The production score and cut-off are unchanged; the overlay is mapped
  back to the same scale by a cut-off table fitted on the adaptation window and alerts at the same
  cut-off $t_b$ (Section~\ref{sec:method:ensemble}).}
\label{fig:framework}
\end{figure*}
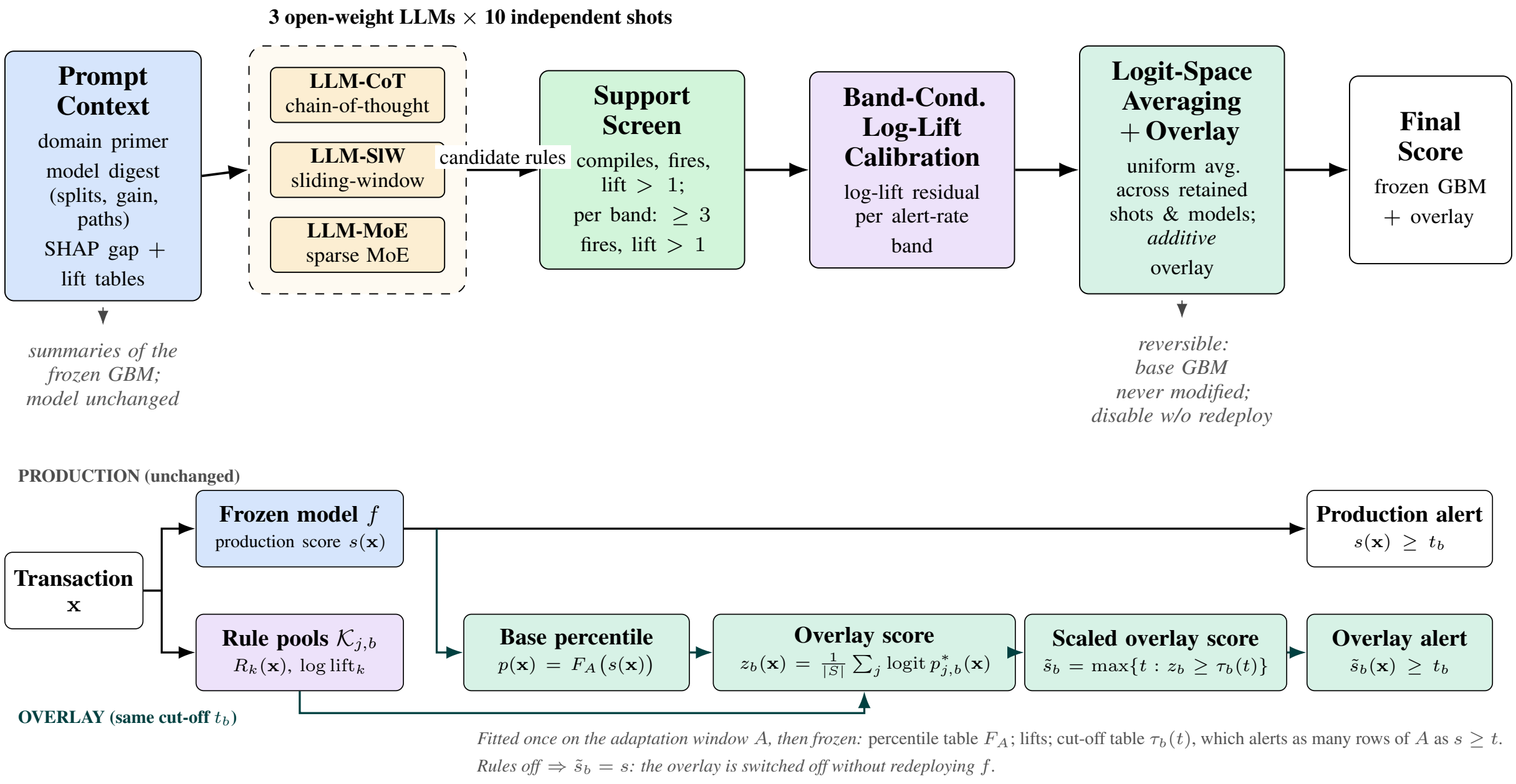

\subsection{Problem Setup and Frozen-Model Constraint}
\label{sec:method:setup}

Let $\mathbf{x}\in\mathcal{X}$ be a transaction feature vector and $y\in\{0,1\}$ the fraud label.
A tree-ensemble base scorer $f$ (constructed in Section~\ref{sec:setup}) is trained once on
historical data and frozen thereafter. Its production score $s(\mathbf{x})\in[0,1000]$ is alerted
at a cut-off $t$, i.e.\ when $s(\mathbf{x})\ge t$. Our framework never modifies $f$, $s$, or $t$.
Because $f$ is never retrained, the generalization-gap failure mode described above is avoided by
construction (Section~\ref{sec:results}).
We call the labeled slice used to generate, screen, and calibrate the overlay the
\emph{adaptation window} $A$; it is not a held-out evaluation set. Every fitted quantity---rules,
lifts, and the lookup tables below---comes from $A$ and is frozen: no out-of-time (OOT) labels or
score distributions are used at any stage. The overlay's base is $p(\mathbf{x})=F_A(s(\mathbf{x}))$,
the score's percentile under $A$'s empirical distribution (a fixed lookup table).

\subsection{Model-Aware LLM Rule Proposal}
\label{sec:method:adv}

For each dataset the framework runs \textbf{3 LLMs $\times$ 10 independent generation shots}: each
shot is a single generation pass with the same prompt (Appendix~\ref{sec:app_prompt}), yielding a
noisy candidate dictionary of diverse predicates.

The prompt has three parts: (i) a domain primer describing the fraud-detection context and feature
encodings; (ii) a digest of the frozen model's trees---per-feature split counts and gain, frequent
feature sets on high-gain paths, and low-value leaves; and (iii) a SHAP attribution gap on the
adaptation window: features are split at the median of their mean signed SHAP over fraud transactions
into those that raise fraud's score on average and those that contribute little or push it toward
legitimate, which we call \emph{underweighted}, with decile (or category) fraud-lift tables for the
latter. The LLM sees these summaries, never the model itself. It reasons about how
fraud would present in the \emph{underweighted} dimensions---behaviours a detector anchored on the
first group would not flag---and every candidate's \emph{primary anchor} must come from that set;
score-raising features may appear only as secondary filters narrowing a pattern to a behavioural
archetype. Each candidate must cite supporting lift from the tables. Candidates are conjunctions of
threshold conditions $(\text{feature}\;\circ\;\theta)$, $\circ\in\{<,\le,>,\ge,=\}$, or missing /
non-missing tests, joined by $\wedge$ (no disjunctions), with thresholds given as training-set
percentiles.

\paragraph{Why an LLM rather than a rule miner?}
The adaptation window's per-band strata are small, so a tree learner or itemset miner fits only
patterns \emph{already present}, while the LLM proposes archetypes from domain priors; its output is
also a named hypothesis with a rationale (Section~\ref{sec:abl:patterns}). We do not compare
empirically against rule-mining baselines and flag this as the primary gap.

\subsection{Empirical Support Screening}
\label{sec:method:screen}

Predicates are compiled with their percentile thresholds taken from the training data; any that fail
to parse or compile are dropped. On the adaptation window, a predicate is kept if it fires, its fraud
rate exceeds the window's base rate (lift $>1$), and it is not a duplicate of an already kept
predicate (canonical form). It then enters the pool $\mathcal{K}_b$ of every band $b$ in whose score
zone $Z_b$ (Section~\ref{sec:method:nb}) it fires on at least three transactions with lift $>1$.
Shots left with no rule in any band are discarded (the \textbf{compilable-shot filter}); retained
shot counts are reported in Section~\ref{sec:setup}. Screening checks empirical support only; it does
not prove semantic correctness or guarantee out-of-time transfer.

\subsection{Band-Conditioned Log-Lift Residual Overlay}
\label{sec:method:nb}

The overlay is fitted for a set of score cut-offs $t_b$ (alert-rate bands $b$). For each band the
framework fits an independent rule pool $\mathcal{K}_b$ and lift set on the adaptation window
\emph{restricted to band $b$'s score zone} $Z_b=\{\mathbf{x}\in A: s(\mathbf{x})\ge t_b\}$.
Let $R_k(\mathbf{x})\in\{0,1\}$ indicate whether rule $k$ fires on $\mathbf{x}$.
For a rule $k\in\mathcal{K}_b$, its calibrated fraud lift is the band-conditioned precision ratio
\begin{equation}
  \mathrm{lift}_k \;=\;
    \frac{P\!\left(y=1 \mid R_k=1,\; \mathbf{x}\in Z_b\right)}
         {P\!\left(y=1 \mid \mathbf{x}\in Z_b\right)},
  \label{eq:lift}
\end{equation}
estimated by a decile-standardized average: $Z_b$ is split into base-score deciles, and the
within-decile precision ratios are averaged weighted by rule-firing support. This controls for
score-dependent base rates and targets---rather than exactly equals---Eq.~\eqref{eq:lift}.
The overlay for cut-off $t_b$ applies $\mathcal{K}_b$ to every transaction:
\begin{equation}
  \boxed{%
  \begin{aligned}
    p^*_b(\mathbf{x}) ={}& \sigma\Bigl(\operatorname{logit}\bigl(p(\mathbf{x})\bigr) \\
    &{}+ \!\!\sum_{k\in\mathcal{K}_b}\!\! \log(\mathrm{lift}_k)\,R_k(\mathbf{x})\Bigr)
  \end{aligned}}
  \label{eq:overlay}
\end{equation}
where $\sigma(\cdot)$ is the sigmoid and $p(\mathbf{x})=F_A(s(\mathbf{x}))$
(Section~\ref{sec:method:setup}). A firing rule with $\mathrm{lift}_k>1$ raises the score, one
below~1 lowers it ($\log$ is natural, matching the logit scale).

\textbf{Zone conditioning.} Lifts are estimated only on the zone where the overlay acts (whole-window
variant: Appendix~\ref{sec:app_pipeabl}); the production replay shares one zone, the lowest cut-off's,
across cut-offs (Section~\ref{sec:setup}).

\subsection{Multi-Shot Averaging, Scaled Score, and Evaluation}
\label{sec:method:ensemble}

Each retained shot $j$ (across all three LLMs) produces $p^*_{j,b}(\mathbf{x})$ via
Eq.~\eqref{eq:overlay}. Stochastic generation produces heterogeneous candidate pools across shots;
the ensemble combines all retained shots $S$ by \textbf{uniform averaging in logit space},
$z_b(\mathbf{x}) = |S|^{-1}\sum_{j\in S}\operatorname{logit}\bigl(p^*_{j,b}(\mathbf{x})\bigr)$.
Every shot receives equal weight and the same combination is applied to all datasets;
logit-space averaging matches the additive parameterization of Eq.~\eqref{eq:overlay} and is a
standard variance-reduction strategy \citep{dietterich2000ensemble}.

\paragraph{Scaled overlay score and deployment.}
To run at the existing cut-off, $z_b$ is mapped back to the production scale by a cut-off table
fitted on $A$: $\tau_b(t)$ is the value at which $z_b$ alerts as many transactions of $A$ as
$s\ge t$, for every $t$ on a 0.1-point grid, and
\begin{equation}
  \tilde s_b(\mathbf{x}) = \max\{\,t : z_b(\mathbf{x}) \ge \tau_b(t)\,\}.
  \label{eq:scaled}
\end{equation}
The overlay alerts when $\tilde s_b(\mathbf{x})\ge t_b$, the base's own cut-off
(Figure~\ref{fig:framework}, bottom). Serving needs only $F_A$, the rules with their lifts, and
$\tau_b$; nothing is refitted on new traffic, and with the rules switched off $\tilde s_b=s$.

\paragraph{Evaluation metric.}
On OOT data we compare production ($s\ge t$) and the overlay ($\tilde s_b\ge t$) through their
precision--recall points over the same threshold grid. Let $P_b$, $R_b$ be production's precision
and recall at its cut-off $t_b$. \textbf{$\Delta$R@P} is the overlay's highest recall at any threshold
with precision $\ge P_b$, minus production's own highest recall at precision $\ge P_b$;
\textbf{$\Delta$P@R} is defined symmetrically at recall $\ge R_b$. Subtracting production's own best
removes what it would gain by merely moving its threshold; no interpolation is used, and a level the
overlay never reaches counts as recall (precision) zero. We report averages over a split's
cut-offs, \textbf{avg$\Delta$R@P} and \textbf{avg$\Delta$P@R}, as absolute differences in \%; the
precision and recall levels they average over are given in Section~\ref{sec:setup}, and every
cut-off in Appendix~\ref{sec:app_perband}. This pair is the sole comparison metric; we do
not use extra-TP counts or PR-AUC/AUC as comparison axes.

\section{Experimental Setup}\label{sec:setup}

\paragraph{Datasets: generalization and robustness to variants.}
We evaluate on two datasets that test the framework's two target capabilities
(Table~\ref{tab:datasets}): generalizing to an unseen population, and staying robust across designed
distribution-shift variants.
\textbf{D-Industrial} detects unauthorized-transaction fraud at a real-world e-commerce platform,
trained on sampled data: training covers approved traffic only, richer in fraud signal but
selection-biased, while evaluation uses a small unbiased cohort approved regardless of score
(Section~\ref{sec:intro}). This is where continued training generalizes worst.
\textbf{BAF}~\citep{jesus2022baf} is a public bank-account benchmark with seven OOT splits of
designed bias-shift variants and no train/evaluation population mismatch: gradient adaptation is
well-posed there, so it tests whether the overlay stays robust when retraining is a viable
alternative. On BAF, screening (Section~\ref{sec:method:screen}) also drops predicates built only
from score-raising features.

\paragraph{Production replay.}
D-Industrial's base is a model we trained on sampled data. We also replay the overlay offline on one of the
platform's legacy production models for the same task, a GBM that, on its own production score
scale, serves as the frozen base; the overlay has not been deployed. Every overlay parameter is fitted on the validation month and frozen for the
next seven (Table~\ref{tab:datasets}), all labelled in October 2026. Shots: LLM-CoT, LLM-SlW (unquantized) and \texttt{Qwen3.6-27B}~\citep{qwen2026qwen36} for LLM-MoE, 10 each, on
in-house GPUs. The model's precision and recall are withheld.

\begin{table*}[t]
  \centering
  \footnotesize
  \setlength{\tabcolsep}{5pt}
  \caption{Datasets and base models, all evaluated out of time; only BAF is public. Blend $\alpha$: LightGBM
    weight in the base's rank blend with a random forest, chosen by validation PR-AUC (sample-weighted on
    D-Industrial); BAF's $\alpha{=}1$ is the sweep's outcome, the replay's base a single model. ---: not disclosed.}
  \label{tab:datasets}
  \begin{tabular}{@{}l p{2.7cm} r r c p{5.3cm}@{}}
    \toprule
    Dataset & Shift type & \#feat & \#trees (LGB\,/\,RF) & Blend $\alpha$ & Split design \\
    \midrule
    D-Industrial
      & selection bias; unbiased eval $\neq$ biased train
      & 500 & 2{,}219\,/\,500 & $0.75$
      & Temporal, six months: train Oct--Nov 2025 (biased, approved-only); validate Dec 2025 and
        evaluate Jan 1--Mar 7, 2026, both from the unbiased cohort \\[2pt]
    BAF
      & public; designed bias-shift variants
      & 30 & 892\,/\,200 & $1.00$
      & Temporal: train months 0--4; evaluate held-out months 5--7 (OOT-1, OOT-2) plus five
        designed bias-shift variants (Var-I--V) \\[2pt]
    Production replay
      & temporal; unbiased eval, seven months
      & --- & --- & $1.00$
      & Temporal: validate Feb 2026; evaluate Mar--Sep 2026 (unbiased cohort) \\
    \bottomrule
  \end{tabular}
\end{table*}

\paragraph{Baselines.}
\textbf{Frozen base} is the scorer under test, never modified; all $\Delta$ metrics are relative to
it. It is itself ensemble-tuned: a rank blend
$\alpha\,\mathrm{rank}(\mathrm{LGB})+(1{-}\alpha)\,\mathrm{rank}(\mathrm{RF})$
\citep{ke2017lightgbm}, ranks taken against the adaptation window, with $\alpha$ swept over a grid
and kept at the value maximizing validation PR-AUC (Table~\ref{tab:datasets}), so overlay gains are
measured over an already ensemble-tuned baseline. \textbf{CT-100} adds 100 boosting rounds on the
adaptation window (a straightforward continued-training baseline). \textbf{CT-Tuned} applies one
boosting round at a lower learning rate (lr$=$0.03), a single round-count/learning-rate sensitivity
point evaluated on D-Industrial only. \textbf{Retrain} fully retrains on the combined
historical-plus-adaptation data. Each is scored on the same scale and evaluated like the overlay,
with its cut-off table fitted on the adaptation window.

\paragraph{Evaluation cut-offs.}
Neither D-Industrial's nor BAF's score scale follows a live alerting policy, so we evaluate at the top
of the 0--1000 scale, where precision is highest: D-Industrial $\{998,\dots,993\}$ (6) and BAF
$\{999,995,990,985,980,975,970\}$ (7). There the frozen base---called production throughout---has
OOT precision and recall of 30.3--56.5\% and 23.5--62.9\% on D-Industrial,
and 10.1--56.2\% and 3.1--50.1\% across the BAF splits; avg$\Delta$R@P and avg$\Delta$P@R average
over these precision and recall levels respectively.

\paragraph{LLM backbones and active shot counts.}
Three architecturally diverse open-weight backbones each generate 10 independent shots per dataset:
\textbf{LLM-CoT} = \texttt{DeepSeek-R1-\allowbreak Distill-Qwen-\allowbreak 32B-4bit}~\citep{deepseek2025r1}
(chain-of-thought); \textbf{LLM-SlW} = \texttt{Gemma4-31B-it-8bit}~\citep{gemma2026gemma4}
(sliding-window attention); \textbf{LLM-MoE} = \texttt{Qwen3.5-35B-A3B-4bit}~\citep{qwen2026qwen35}
(mixture-of-experts). Shots left with no rule after screening are discarded; active counts after
filtering (LLM-CoT\,/\,LLM-SlW\,/\,LLM-MoE) are D-Industrial \mbox{9\,/\,8\,/\,8} and BAF
\mbox{10\,/\,6\,/\,9} (19/14/17 total). The all-3 ensemble averages uniformly over all retained
shots. Generation ran locally on one Apple M4 Max workstation (128\,GB unified memory); no hosted
API was used, so transaction data never leaves the environment.

\section{Results}\label{sec:results}

\subsection{Continued Training vs.\ the Overlay}\label{sec:results:conttrain}

We compare the all-3 ensemble overlay with CT-Tuned, CT-100 and Retrain (full figures in
Appendix~\ref{sec:app_conttrain}, Table~\ref{tab:conttrain}).

\begin{table*}[t]
  \centering
  \footnotesize
  \renewcommand{\arraystretch}{0.95}
  \caption{Per-backbone and ensemble results: \textbf{avg$\Delta$R@P} / avg$\Delta$P@R (absolute
    differences, \%) relative to the frozen base at its evaluation cut-offs. \emph{Anchors}: production's
    OOT precision and recall there, the levels at which $\Delta$R@P and $\Delta$P@R are taken. \textbf{Ensemble} is the primary result (bold where
    positive). Var-IV shares Var-III's population. CoT, SlW, MoE: Section~\ref{sec:setup}.}
  \label{tab:main}
  \resizebox{\textwidth}{!}{%
  \begin{tabular}{@{}ll cc rrrr@{}}
    \toprule
    & & \multicolumn{2}{c}{Anchors (production)} & & & & \\
    \cmidrule(lr){3-4}
    Dataset & Split & Precision & Recall & CoT & SlW & MoE & \textbf{Ensemble (All 3)} \\
    \midrule
    D-Industrial & OOT & 30.3--56.5\% & 23.5--62.9\% & $-0.28\%$ / $-0.15\%$ & $+2.27\%$ / $+1.56\%$ & $+3.88\%$ / $+3.09\%$ & $\mathbf{+2.26\%}$ / $\mathbf{+1.60\%}$ \\
    \midrule
    BAF & OOT-1 & 17.3--53.2\% & 4.1--41.4\% & $-0.12\%$ / $+0.12\%$ & $-0.21\%$ / $+0.06\%$ & $-0.18\%$ / $-0.09\%$ & $-0.10\%$ / $\mathbf{+0.07\%}$ \\
     & OOT-2 & 19.1--52.8\% & 3.9--41.2\% & $-0.08\%$ / $+0.09\%$ & $-0.41\%$ / $-0.17\%$ & $-0.26\%$ / $-0.14\%$ & $0.00\%$ / $\mathbf{+0.06\%}$ \\
     & Var-I & 18.0--56.2\% & 3.1--35.3\% & $+0.21\%$ / $+0.12\%$ & $-0.07\%$ / $+0.17\%$ & $-0.22\%$ / $+0.04\%$ & $\mathbf{+0.04\%}$ / $\mathbf{+0.08\%}$ \\
     & Var-II & 14.6--51.8\% & 6.4--50.1\% & $+0.37\%$ / $+0.25\%$ & $+0.20\%$ / $+0.13\%$ & $+0.19\%$ / $+0.13\%$ & $\mathbf{+0.27\%}$ / $\mathbf{+0.30\%}$ \\
     & Var-III & 10.1--36.6\% & 4.5--43.1\% & $+0.23\%$ / $+0.26\%$ & $-0.22\%$ / $+0.18\%$ & $-0.07\%$ / $+0.41\%$ & $\mathbf{+0.04\%}$ / $\mathbf{+0.28\%}$ \\
     & Var-IV & 10.1--36.6\% & 4.5--43.1\% & $+0.23\%$ / $+0.26\%$ & $-0.22\%$ / $+0.18\%$ & $-0.07\%$ / $+0.41\%$ & $\mathbf{+0.04\%}$ / $\mathbf{+0.28\%}$ \\
     & Var-V & 10.8--40.4\% & 5.1--45.6\% & $+0.10\%$ / $+0.01\%$ & $-0.51\%$ / $-0.13\%$ & $-0.36\%$ / $-0.28\%$ & $-0.13\%$ / $\mathbf{+0.02\%}$ \\
    \bottomrule
  \end{tabular}%
  }
\end{table*}

\paragraph{D-Industrial: the generalization test.}
The \emph{fair} comparator here is a carefully tuned continued-training update, not a naive one. Even
CT-Tuned (one boosting round, lr$=$0.03) regresses, at avg$\Delta$R@P $-7.69\%$ and avg$\Delta$P@R
$-2.85\%$; the naive CT-100 and full Retrain never reach production's precision at any cut-off (avg$\Delta$R@P $-46.39\%$: no recall at those precision levels) and lose $34.80\%$ and $41.73\%$ of precision at equal recall. The overlay instead
\emph{gains} avg$\Delta$R@P $+2.26\%$ (95\% CI $[+1.84\%,+2.90\%]$, paired block bootstrap
over 66 calendar-day blocks, 1000 replicates) and avg$\Delta$P@R $+1.60\%$
($[+1.28\%,+2.04\%]$), positive at every cut-off over the OOT period and on average in each month
(Appendix~\ref{sec:app_month}), with no redeployment. The cause
is structural, not under-tuning: every training example carries the same selection bias, so no
learning rate or round count turns the biased gradient direction into an unbiased
one~\citep{quinonero2009shift}; a frozen base sidesteps the mechanism entirely.

\paragraph{BAF: a robustness floor.}
With no population gap, gradient updates are well-posed and the overlay should not be expected to
lead---and it does not. It does no harm: avg$\Delta$P@R is positive on all seven splits
($[+0.02\%,+0.30\%]$) and avg$\Delta$R@P stays near neutral ($[-0.13\%,+0.27\%]$;
negative on OOT-1 and Var-V). CT-100 is no better ($[-0.39\%,-0.04\%]$ / $[-0.29\%,+0.32\%]$); Retrain raises precision at equal recall more ($[+0.08\%,+0.71\%]$) but loses recall at equal precision on six splits (down to $-0.75\%$), and both need a redeploy.

\begin{table}[t]
  \centering
  \footnotesize
  \setlength{\tabcolsep}{3pt}
  \caption{Production replay by OOT month, averaged over the high-precision cut-offs of most business interest;
    95\% paired day-bootstrap intervals; $^*$: excludes 0.}
  \label{tab:replay}
  \begin{tabular}{@{}lrr@{}}
    \toprule
    Month & avg$\Delta$R@P & avg$\Delta$P@R \\
    \midrule
    Mar & $+15.89\%^*$\,{\scriptsize$[+6.1,+20.0]$} & $+3.87\%^*$\,{\scriptsize$[+2.2,+6.1]$} \\
    Apr & $+20.66\%\phantom{^*}$\,{\scriptsize$[-0.7,+23.0]$} & $+4.86\%^*$\,{\scriptsize$[+2.3,+7.3]$} \\
    May & $+3.84\%\phantom{^*}$\,{\scriptsize$[-4.7,+9.1]$} & $+2.89\%\phantom{^*}$\,{\scriptsize$[-5.4,+13.1]$} \\
    Jun & $+6.25\%\phantom{^*}$\,{\scriptsize$[-3.8,+8.0]$} & $+2.37\%\phantom{^*}$\,{\scriptsize$[-1.9,+7.7]$} \\
    Jul & $+5.35\%^*$\,{\scriptsize$[+1.4,+7.3]$} & $+5.00\%^*$\,{\scriptsize$[+1.7,+7.6]$} \\
    Aug & $+6.06\%^*$\,{\scriptsize$[+0.9,+10.9]$} & $+6.05\%^*$\,{\scriptsize$[+2.0,+9.6]$} \\
    Sep & $+5.12\%\phantom{^*}$\,{\scriptsize$[-0.4,+8.1]$} & $+8.81\%^*$\,{\scriptsize$[+2.9,+16.7]$} \\
    \midrule
    Mar--Sep & $+13.18\%^*$\,{\scriptsize$[+9.6,+14.6]$} & $+5.64\%^*$\,{\scriptsize$[+4.3,+7.1]$} \\
    \bottomrule
  \end{tabular}
\end{table}

\subsection{Production Replay}\label{sec:results:replay}

Fitted on one month and frozen for seven (Table~\ref{tab:replay}), the overlay is more precise at
equal recall than the legacy model in every month, the latest included: avg$\Delta$P@R is
positive throughout, its interval excludes zero in five of seven months, and it is $+5.64\%$ pooled
($[+4.34\%,+7.14\%]$). Recall at equal precision gains more ($+13.18\%$) because production's
precision is nearly flat over these cut-offs, so a small precision gain moves the equal-precision
point far down the curve. Adding lower-precision cut-offs keeps the pooled gains positive
($+6.47\%$ / $+2.61\%$), though there the overlay trails production slightly ($-0.91\%$ / $-0.72\%$).

\subsection{Proposal-Source Variance and Ensemble Stabilization}\label{sec:results:main}

Table~\ref{tab:main} reports both metrics per backbone and for the all-3 ensemble. The central
finding is \emph{high single-LLM variance}: the regimes invert the ranking---LLM-MoE is best
industrially ($+3.88\%$ avg$\Delta$R@P) yet negative on six of seven BAF splits, while
LLM-CoT is slightly negative on D-Industrial ($-0.28\%$) yet the best single source on
\emph{all seven}. No in-distribution signal separates them, and their rules barely overlap
(Appendix~\ref{sec:app_srcabl}).

Ensembling all three resolves this without an oracle: avg$\Delta$P@R is positive on \emph{all eight}
splits and avg$\Delta$R@P on five (the exceptions within $0.13\%$ of neutral), trading part
of LLM-MoE's industrial peak ($+2.26\%$ vs $+3.88\%$) to avoid its BAF degradation.
The gain also needs the rest of the pipeline (Appendix~\ref{sec:app_pipeabl}): without the log-lift
overlay or the per-shot ensemble the same LLM rules fall below the frozen base on every split.

\paragraph{Readable rules.}\label{sec:abl:patterns}
The LLMs largely follow the steering instruction: the first condition of 97\% (D-Industrial) and 99\% (BAF)
of retained predicates is an underweighted feature. On D-Industrial, LLM-CoT proposes
\texttt{\scriptsize failed\_txn(addr,14d)>p90 $\wedge$ buyer\_seg>p80}, reasoning that repeated failed
payments to one address test card authorization; on the adaptation window it reaches $71.6\times$
lift on 6{,}249 transactions.

\section{Conclusion}\label{sec:conclusion}

LLMs reasoning over a frozen fraud detector's diagnostics patched its blind spots: their rules,
steered through its underweighted dimensions, form a reversible log-lift overlay at the production
cut-off. It
generalizes where continued training regresses, holds for seven months on a legacy production model,
stays robust
where retraining offers no edge, and keeps each amendment auditable; pooling
backbones bounds the worst case. Beyond predicting, LLMs
can reason about where models fail.

\section*{Limitations}

\paragraph{Scope of the recall advantage.}
The recall advantage is demonstrated under a generalization gap between training and evaluation
populations. On BAF, where no such gap exists, neither method dominates on recall and we claim only
that the overlay does no harm.

\paragraph{Continued-training baselines.}
We evaluate two continued-training configurations on D-Industrial (CT-100 and CT-Tuned); both trail
the frozen baseline, but two configurations do not span the hyperparameter space. Importance-weighted
or propensity-adjusted continued training is the principled remedy; we do not evaluate it, as the
selection mechanism is a separate model whose scores are unavailable to the adaptation procedure.
The production replay compares against the legacy model itself only, without these baselines.

\paragraph{Inference cost.}
The three-model ensemble requires roughly $3\times$ the LLM inference budget of a single-source
overlay; a single-source fallback is feasible but sacrifices variance reduction.

\paragraph{Backbone scale and model context.}
All backbones are open-weight models of 27--35B parameters run in-house, and the prompt summarizes
the frozen model rather than exposing it: a digest of its highest-gain features and paths, SHAP-gap
tables, and thresholds restricted to a fixed grid of training-set percentiles. We did not test larger
backbones or richer context, such as finer thresholds or more of the tree structure, which might
propose rules the current summaries cannot express.

\paragraph{Offline evaluation.}
All results are offline, with every parameter fixed on the adaptation window. On both industrial
settings, evaluation uses the unbiased cohort---the platform's universal control group, approved
regardless of score---so fraud labels are free of the selection bias that live blocking introduces, much as in an
A/B holdout. Offline evaluation cannot capture the feedback a deployed overlay would trigger, such as
fraudsters adapting to its rules; the overlay has not been deployed. The replay's third backbone
differs from the main experiments' (Section~\ref{sec:setup}).

\section*{Ethics Statement}

\paragraph{Data and privacy.}
Experiments on D-Industrial and the production replay use historical transaction records accessed
under the platform's
internal data-governance program. No human-subject data were collected, no directly identifying
attributes are used as model features, and feature names are paraphrased throughout. BAF is a public
benchmark released for fraud-detection research.

\paragraph{Risk of harm.}
Fraud decisions carry asymmetric costs: a false alert burdens a legitimate customer, a missed fraud
harms the platform and its users. The overlay alerts at the production cut-off and never blocks
autonomously; deployments should retain human review and an appeals path.

\bibliography{custom}

\appendix

\section{Continued-Training Comparison}
\label{sec:app_conttrain}

Table~\ref{tab:conttrain} gives the full continued-training comparison summarized in
Section~\ref{sec:results:conttrain}: the all-3 ensemble overlay against three continued-training
variants (CT-100, CT-Tuned, full Retrain) under the same avg$\Delta$R@P / avg$\Delta$P@R metric,
across both datasets. Each method's score is mapped to the production scale by its own cut-off table
fitted on the adaptation window (Section~\ref{sec:method:ensemble}).

\begin{table*}[t]
  \centering
  \footnotesize
  \setlength{\tabcolsep}{3pt}
  \caption{Comparison with continued training: \textbf{avg$\Delta$R@P} / avg$\Delta$P@R (\%)
    relative to the frozen base at its evaluation cut-offs. BAF figures are ranges over the 7 splits.
    On D-Industrial, CT-100 and Retrain never reach production's precision at any cut-off, so
    their recall at those precision levels is zero. CT-Tuned (1 round, lr$=$0.03) evaluated on
    D-Industrial only. \textbf{Bold column} = our framework (all-3 ensemble).}
  \label{tab:conttrain}
  \resizebox{\textwidth}{!}{%
  \begin{tabular}{@{}l rr rr rr rr@{}}
    \toprule
    & \multicolumn{2}{c}{CT-100} & \multicolumn{2}{c}{CT-Tuned} & \multicolumn{2}{c}{Retrain}
    & \multicolumn{2}{c}{\textbf{Ours}} \\
    \cmidrule(lr){2-3}\cmidrule(lr){4-5}\cmidrule(lr){6-7}\cmidrule(lr){8-9}
    Dataset (split) & \textbf{$\Delta$R@P} & $\Delta$P@R
                    & \textbf{$\Delta$R@P} & $\Delta$P@R
                    & \textbf{$\Delta$R@P} & $\Delta$P@R
                    & \textbf{$\Delta$R@P} & $\Delta$P@R \\
    \midrule
    D-Industrial OOT
      & $-46.39\%$ & $-34.80\%$
      & $-7.69\%$ & $-2.85\%$
      & $-46.39\%$ & $-41.73\%$
      & $\mathbf{+2.26\%}$ & $\mathbf{+1.60\%}$ \\
    BAF (7 splits, range)
      & $[-0.39\%,-0.04\%]$ & $[-0.29\%,+0.32\%]$
      & \multicolumn{2}{c}{---}
      & $[-0.75\%,+0.09\%]$ & $[+0.08\%,+0.71\%]$
      & $[-0.13\%,+0.27\%]$ & $[+0.02\%,+0.30\%]$ \\
    \bottomrule
  \end{tabular}%
  }
\end{table*}

\section{Prompt Design}
\label{sec:app_prompt}

Each generation shot sends a single prompt to the LLM; there is no second LLM pass.
Listing~\ref{lst:prompt} reproduces the template behind all reported results (D-Industrial and BAF,
all three backbones), as recovered from the logged LLM calls. Four placeholders are filled per
dataset:
\begin{itemize}[nosep,leftmargin=1.1em,topsep=2pt]
  \item \texttt{\{domain primer\}}: a description of the fraud-detection task and of the dataset's
    features and their encodings. It is dataset-specific and not reproduced; for D-Industrial it
    describes proprietary data.
  \item \texttt{\{model digest\}}: a summary of the frozen model's trees---split counts, total gain and
    split depths for the features with the most gain, the names of the remaining features, the
    feature sets most frequent on high-gain leaf paths, and a few low-value leaves.
  \item \texttt{\{SHAP gap\}}: features split at the median of their mean signed SHAP over fraud transactions
    on the adaptation window, into those above it (``forbidden'', listed with their mean SHAP) and
    those at or below it (``target''), each target feature with a decile or category table of counts,
    fraud rate, and lift over the window's base rate. The template's ``SHAP \texttt{<=} 0'' refers to
    this target set.
  \item \texttt{\{number of rules\}}: 50.
\end{itemize}
All other text is verbatim, with non-ASCII glyphs rendered in ASCII (an em dash as \texttt{-{}-},
$\leq$ as \texttt{<=}, and a left-pointing marker as \texttt{<-}). The returned predicates are then
compiled, screened and assigned to band pools in code (Section~\ref{sec:method:screen}), not by the
LLM.

\begin{lstlisting}[basicstyle=\ttfamily\scriptsize,breaklines=true,columns=fullflexible,keepspaces=true,numbers=none,frame=single,captionpos=b,caption={Rule-proposal prompt template used for every shot (anonymized placeholders, ASCII).},label={lst:prompt}]
{domain primer}

---

{model digest}

---

{SHAP gap}

---

## Task: Adversarial Rule Generation

You are a red-team analyst. The LightGBM model already captures fraud patterns
anchored on its **forbidden features** (high SHAP for fraud, listed above).
Your job is to find fraud patterns the model systematically misses by reasoning
in the **target feature** dimensions it underweights.

Work through the following three steps in your head before writing the output.

### Step 1 -- Adversarial evasion reasoning

Think like a fraudster who knows LGB monitors the forbidden features closely.
Design fraud behaviours that primarily exploit the target features
(SHAP <= 0, listed above) as the main discriminating anchor.
You may also include forbidden features as secondary filters when the combination
reveals a behavioural archetype that the target features alone would not isolate --
for example, a target-feature profile that only becomes high-precision when
narrowed by a forbidden-feature threshold.
For each candidate pattern, ask: *why would a fraudster naturally exhibit this
combination? Does it correspond to a real behavioural archetype?*

### Step 2 -- Data grounding

For each evasion pattern, check the decile/category tables in the SHAP Gap
section. Look for target-feature groups with **Lift > 1.5 (<-)**. Prefer
patterns that are both theoretically coherent AND supported by elevated lift
in the data. If no <- appears for a candidate feature, weigh it lower.

### Step 3 -- Predicate construction

Translate the evasion pattern into a conjunction predicate using:
- Feature names exactly as listed above (no invented names).
- Operators: `>`, `<`, `>=`, `<=`, `==`, `isna`, `notna`.
- Percentile tokens: `p5`, `p10`, `p15`, `p20`, `p25`, `p30`, `p40`,
  `p50`, `p60`, `p70`, `p75`, `p80`, `p85`, `p90`, `p95`, `p99`.
- Join conditions with `AND` only (no top-level OR).

---

### Output format

Produce exactly **{number of rules}** rules as a JSON array. Each item:

```json
[
  {
    "id": "R_001",
    "name": "short label (5 words max)",
    "evasion_reasoning": "why a fraudster shows this pattern in target features; if forbidden features appear as secondary filters, explain why that narrows to a distinct archetype",
    "data_evidence": "specific lift values or decile ranges from the table that support this",
    "target_features_used": ["feat_a", "feat_b"],
    "forbidden_features_used": ["feat_c"],
    "predicate": "feat_a > p80 AND feat_b == 3 AND feat_c > p90"
  }
]
```

Constraints:
- Use ONLY feature names that appear verbatim in the digest or SHAP gap above.
- The primary anchor condition(s) must be target features (SHAP <= 0). Forbidden features may appear as secondary filters in a conjunction.
- Each rule must address a distinct evasion archetype -- no near-duplicates.
- Rank from strongest evidence + clearest evasion logic to weakest.

Respond with the JSON array only -- no explanation, no markdown fences outside the JSON.
\end{lstlisting}

\section{Full Pipeline Ablation}
\label{sec:app_pipeabl}

Table~\ref{tab:pipeabl_full} reports both metrics across all eight evaluation splits.
Conditions denote removing components from \textbf{Full} (complete pipeline: all three backbones,
per-band calibration, per-shot logit ensemble): $-$PerBand replaces per-band with global (whole-window)
calibration; $-$Ens uses a merged deduplicated rule pool with no per-shot ensemble;
$-$Ens\,$-$PerBand combines both; $-$LogLift scores each shot by its rule fire rate instead of the
log-lift overlay; Rules-only uses the merged pool's fire rate. Every condition is evaluated like
Full, with its own cut-off table fitted on the adaptation window.

\begin{table*}[t]
\centering
\footnotesize
\caption{Full pipeline ablation: avg$\Delta$R@P (top) and avg$\Delta$P@R (bottom), \%, per split.
  \textbf{Full} = complete pipeline; each other column removes components
  (definitions in Appendix~\ref{sec:app_pipeabl}).}
\label{tab:pipeabl_full}
\begin{tabular}{llrrrrrr}
\toprule
Dataset & Split   & \textbf{Full} & $-$PerBand & $-$Ens & $-$Ens\,$-$PerBand & $-$LogLift & Rules-only \\
\midrule
\multicolumn{8}{@{}l}{\textit{avg$\Delta$R@P}} \\
D-Industrial & OOT & $\mathbf{+2.26\%}$ & $-2.41\%$ & $-2.48\%$ & $-46.39\%$ & $-35.73\%$ & $-24.76\%$ \\
BAF & OOT-1 & $-0.10\%$ & $-0.42\%$ & $-21.42\%$ & $-22.87\%$ & $-26.02\%$ & $-25.76\%$ \\
BAF & OOT-2 & $0.00\%$ & $-0.38\%$ & $-20.44\%$ & $-20.97\%$ & $-25.46\%$ & $-25.16\%$ \\
BAF & Var-I & $\mathbf{+0.04\%}$ & $-0.09\%$ & $-17.45\%$ & $-19.12\%$ & $-21.46\%$ & $-21.04\%$ \\
BAF & Var-II & $\mathbf{+0.27\%}$ & $-0.06\%$ & $-19.96\%$ & $-24.10\%$ & $-32.92\%$ & $-32.10\%$ \\
BAF & Var-III & $\mathbf{+0.04\%}$ & $-0.20\%$ & $-17.85\%$ & $-21.41\%$ & $-26.79\%$ & $-26.40\%$ \\
BAF & Var-IV & $\mathbf{+0.04\%}$ & $-0.20\%$ & $-17.85\%$ & $-21.41\%$ & $-26.79\%$ & $-26.40\%$ \\
BAF & Var-V & $-0.13\%$ & $-0.70\%$ & $-21.97\%$ & $-24.63\%$ & $-29.18\%$ & $-28.44\%$ \\
\midrule
\multicolumn{8}{@{}l}{\textit{avg$\Delta$P@R}} \\
D-Industrial & OOT & $\mathbf{+1.60\%}$ & $-2.99\%$ & $-5.11\%$ & $-41.54\%$ & $-31.45\%$ & $-24.11\%$ \\
BAF & OOT-1 & $\mathbf{+0.07\%}$ & $-0.25\%$ & $-15.46\%$ & $-19.38\%$ & $-26.84\%$ & $-26.00\%$ \\
BAF & OOT-2 & $\mathbf{+0.06\%}$ & $-0.21\%$ & $-16.18\%$ & $-19.68\%$ & $-28.15\%$ & $-27.15\%$ \\
BAF & Var-I & $\mathbf{+0.08\%}$ & $+0.25\%$ & $-15.73\%$ & $-17.79\%$ & $-26.75\%$ & $-26.37\%$ \\
BAF & Var-II & $\mathbf{+0.30\%}$ & $+0.10\%$ & $-13.45\%$ & $-16.36\%$ & $-23.78\%$ & $-22.71\%$ \\
BAF & Var-III & $\mathbf{+0.28\%}$ & $+0.26\%$ & $-9.25\%$ & $-11.49\%$ & $-16.42\%$ & $-15.69\%$ \\
BAF & Var-IV & $\mathbf{+0.28\%}$ & $+0.26\%$ & $-9.25\%$ & $-11.49\%$ & $-16.42\%$ & $-15.69\%$ \\
BAF & Var-V & $\mathbf{+0.02\%}$ & $+0.27\%$ & $-11.20\%$ & $-12.10\%$ & $-17.61\%$ & $-16.74\%$ \\
\bottomrule
\end{tabular}
\end{table*}

\section{LLM Sources: Ablation, Proposal Behaviour and Patterns}
\label{sec:app_srcabl}

Table~\ref{tab:srcabl_full} reports both metrics across all eight evaluation splits for all seven
combinations of the three LLM backbones (full pipeline). No combination dominates uniformly:
LLM-MoE alone gives the highest avg$\Delta$R@P on D-Industrial ($+3.88\%$) and
LLM-SlW$+$LLM-MoE the highest of the pairs ($+2.80\%$), but both are negative on most BAF
splits, while LLM-CoT---slightly negative on D-Industrial---is the best single source on every BAF
split. All~3 remains strongly positive on D-Industrial and avoids the worst corners on BAF.

The backbones also differ in how they follow the output contract (Appendix~\ref{sec:app_prompt}). On
D-Industrial, 44\% of LLM-CoT's returned predicates name features that do not exist in the
model---half of them near-misses of real names---against 2\% or less for the other two, and
62\% of LLM-SlW's are repeats within a shot; on both datasets, 69--73\% of LLM-MoE's pass the support
screen. The retained rule sets barely overlap: only 14\% (D-Industrial) and 16\% (BAF) of distinct
retained predicates come from more than one shot, and LLM-CoT shares just 2 of its 22 D-Industrial
rules with each other backbone. Each backbone thus contributes largely different patterns---the
variance the ensemble absorbs.

\paragraph{Representative patterns.}
Each retained predicate carries the LLM's name and rationale beside its measured lift. Besides an
underweighted first condition (97\% on D-Industrial, 99\% on BAF), score-raising features appear as
secondary filters in 9\% and 33\% of retained predicates. Rules typically combine individually weak
evidence into a selective conjunction: in the D-Industrial example of Section~\ref{sec:abl:patterns},
LLM-CoT cites only a $1.09\times$ lift for failed payments at one shipping address. On BAF, LLM-MoE
combines \texttt{\scriptsize session\_len>p90} ($1.44\times$) and \texttt{\scriptsize free\_email}
($1.22\times$) with \texttt{\scriptsize housing=0} into a $6.7\times$ rule (862 transactions).
Calibration lift does not predict which generalize: the highest-lift D-Industrial predicates come
from LLM-CoT, the weakest single source there.

\begin{table*}[t]
\centering
\footnotesize
\setlength{\tabcolsep}{4pt}
\caption{Full LLM-source ablation: avg$\Delta$R@P (top) and avg$\Delta$P@R (bottom), \%, per split,
  all seven backbone combinations (full pipeline). \textbf{All~3} = our default ensemble.
  CoT/SlW/MoE abbreviate LLM-CoT/LLM-SlW/LLM-MoE (Section~\ref{sec:setup}).}
\label{tab:srcabl_full}
\begin{tabular}{llrrrrrrr}
\toprule
Dataset & Split & CoT & SlW & MoE & CoT+SlW & CoT+MoE & SlW+MoE & \textbf{All~3} \\
\midrule
\multicolumn{9}{@{}l}{\textit{avg$\Delta$R@P}} \\
D-Industrial & OOT & $-0.28\%$ & $+2.27\%$ & $+3.88\%$ & $+1.14\%$ & $+2.36\%$ & $+2.80\%$ & $\mathbf{+2.26\%}$ \\
BAF & OOT-1 & $-0.12\%$ & $-0.21\%$ & $-0.18\%$ & $+0.06\%$ & $-0.03\%$ & $-0.33\%$ & $-0.10\%$ \\
BAF & OOT-2 & $-0.08\%$ & $-0.41\%$ & $-0.26\%$ & $+0.14\%$ & $+0.08\%$ & $-0.29\%$ & $0.00\%$ \\
BAF & Var-I & $+0.21\%$ & $-0.07\%$ & $-0.22\%$ & $+0.27\%$ & $+0.16\%$ & $-0.20\%$ & $\mathbf{+0.04\%}$ \\
BAF & Var-II & $+0.37\%$ & $+0.20\%$ & $+0.19\%$ & $+0.27\%$ & $+0.33\%$ & $+0.18\%$ & $\mathbf{+0.27\%}$ \\
BAF & Var-III & $+0.23\%$ & $-0.22\%$ & $-0.07\%$ & $+0.27\%$ & $+0.26\%$ & $-0.18\%$ & $\mathbf{+0.04\%}$ \\
BAF & Var-IV & $+0.23\%$ & $-0.22\%$ & $-0.07\%$ & $+0.27\%$ & $+0.26\%$ & $-0.18\%$ & $\mathbf{+0.04\%}$ \\
BAF & Var-V & $+0.10\%$ & $-0.51\%$ & $-0.36\%$ & $+0.02\%$ & $-0.12\%$ & $-0.42\%$ & $-0.13\%$ \\
\midrule
\multicolumn{9}{@{}l}{\textit{avg$\Delta$P@R}} \\
D-Industrial & OOT & $-0.15\%$ & $+1.56\%$ & $+3.09\%$ & $+0.78\%$ & $+1.80\%$ & $+2.34\%$ & $\mathbf{+1.60\%}$ \\
BAF & OOT-1 & $+0.12\%$ & $+0.06\%$ & $-0.09\%$ & $+0.27\%$ & $+0.06\%$ & $-0.05\%$ & $\mathbf{+0.07\%}$ \\
BAF & OOT-2 & $+0.09\%$ & $-0.17\%$ & $-0.14\%$ & $+0.24\%$ & $+0.06\%$ & $-0.04\%$ & $\mathbf{+0.06\%}$ \\
BAF & Var-I & $+0.12\%$ & $+0.17\%$ & $+0.04\%$ & $+0.19\%$ & $+0.13\%$ & $+0.44\%$ & $\mathbf{+0.08\%}$ \\
BAF & Var-II & $+0.25\%$ & $+0.13\%$ & $+0.13\%$ & $+0.23\%$ & $+0.16\%$ & $+0.11\%$ & $\mathbf{+0.30\%}$ \\
BAF & Var-III & $+0.26\%$ & $+0.18\%$ & $+0.41\%$ & $+0.25\%$ & $+0.33\%$ & $+0.37\%$ & $\mathbf{+0.28\%}$ \\
BAF & Var-IV & $+0.26\%$ & $+0.18\%$ & $+0.41\%$ & $+0.25\%$ & $+0.33\%$ & $+0.37\%$ & $\mathbf{+0.28\%}$ \\
BAF & Var-V & $+0.01\%$ & $-0.13\%$ & $-0.28\%$ & $+0.09\%$ & $+0.09\%$ & $+0.01\%$ & $\mathbf{+0.02\%}$ \\
\bottomrule
\end{tabular}
\end{table*}

\section{All Evaluation Cut-offs}
\label{sec:app_perband}

Table~\ref{tab:perband_all} lists, for every cut-off $t_b$ and split, production ($s\ge t_b$) and
the all-3 ensemble overlay ($\tilde s_b\ge t_b$) side by side on OOT data---precision and recall---with
the $\Delta$R@P and $\Delta$P@R checks at that cut-off. The two need not alert the same volume on
OOT; the checks compare them at equal precision and equal recall instead. Var-IV shares Var-III's evaluation population.

\begin{table*}[t]
\centering
\scriptsize
\setlength{\tabcolsep}{4pt}
\caption{All evaluation cut-offs, all-3 ensemble: production $s\ge t_b$ vs.\ overlay
  $\tilde s_b\ge t_b$ on OOT data (precision P\%, recall R\%), and $\Delta$R@P / $\Delta$P@R
  (\%) at each cut-off. Every parameter is fixed on the adaptation window.}
\label{tab:perband_all}
\begin{tabular}{llrrrrrr}
\toprule
& & \multicolumn{2}{c}{Production} & \multicolumn{2}{c}{Overlay} & & \\
\cmidrule(lr){3-4}\cmidrule(lr){5-6}
Split & $t_b$ & P\% & R\% & P\% & R\% & $\Delta$R@P & $\Delta$P@R \\
\midrule
D-Ind. OOT & 998 & 56.48 & 23.47 & 57.51 & 24.37 & $+2.57\%$ & $+1.03\%$ \\
 & 997 & 49.07 & 35.67 & 50.30 & 38.00 & $+3.31\%$ & $+2.04\%$ \\
 & 996 & 43.29 & 45.80 & 44.06 & 47.67 & $+1.88\%$ & $+1.66\%$ \\
 & 995 & 37.99 & 52.41 & 38.73 & 54.00 & $+2.41\%$ & $+1.91\%$ \\
 & 994 & 33.99 & 58.12 & 34.58 & 59.27 & $+1.55\%$ & $+1.96\%$ \\
 & 993 & 30.32 & 62.86 & 30.91 & 64.00 & $+1.84\%$ & $+1.02\%$ \\
\midrule
BAF OOT-1 & 999 & 53.15 & 4.12 & 53.95 & 4.05 & $+0.21\%$ & $+0.30\%$ \\
 & 995 & 34.89 & 13.88 & 34.68 & 13.53 & $-0.59\%$ & $-0.11\%$ \\
 & 990 & 28.15 & 22.40 & 28.55 & 21.74 & $-0.31\%$ & $-0.34\%$ \\
 & 985 & 23.76 & 28.35 & 24.39 & 28.00 & $+0.94\%$ & $+0.52\%$ \\
 & 980 & 20.77 & 33.07 & 21.08 & 33.31 & $+0.94\%$ & $+0.60\%$ \\
 & 975 & 18.94 & 37.68 & 19.00 & 37.05 & $-1.68\%$ & $-0.45\%$ \\
 & 970 & 17.35 & 41.42 & 17.56 & 40.96 & $-0.21\%$ & $-0.01\%$ \\
\midrule
BAF OOT-2 & 999 & 52.83 & 3.89 & 52.68 & 3.75 & $+0.31\%$ & $+0.71\%$ \\
 & 995 & 37.20 & 13.83 & 36.60 & 13.24 & $-1.53\%$ & $-0.47\%$ \\
 & 990 & 29.74 & 21.86 & 30.08 & 21.09 & $-0.35\%$ & $-0.22\%$ \\
 & 985 & 25.23 & 27.52 & 25.54 & 26.93 & $+1.08\%$ & $+0.36\%$ \\
 & 980 & 22.57 & 32.97 & 22.64 & 32.80 & $+0.69\%$ & $+0.10\%$ \\
 & 975 & 20.51 & 37.32 & 20.70 & 36.80 & $-0.14\%$ & $-0.01\%$ \\
 & 970 & 19.07 & 41.17 & 19.30 & 40.93 & $-0.10\%$ & $-0.08\%$ \\
\midrule
BAF Var-I & 999 & 56.25 & 3.13 & 57.79 & 3.09 & $-0.03\%$ & $-0.25\%$ \\
 & 995 & 33.98 & 10.88 & 34.50 & 10.63 & $+0.07\%$ & $+0.07\%$ \\
 & 990 & 28.05 & 17.72 & 28.91 & 17.48 & $0.00\%$ & $+0.56\%$ \\
 & 985 & 24.64 & 23.45 & 25.22 & 23.11 & $+0.59\%$ & $+0.39\%$ \\
 & 980 & 21.85 & 28.11 & 22.23 & 28.04 & $+0.56\%$ & $+0.30\%$ \\
 & 975 & 19.65 & 32.00 & 19.97 & 31.62 & $-0.28\%$ & $-0.17\%$ \\
 & 970 & 17.99 & 35.27 & 18.03 & 34.64 & $-0.63\%$ & $-0.33\%$ \\
\midrule
BAF Var-II & 999 & 51.84 & 6.36 & 53.04 & 6.36 & $-0.21\%$ & $+1.18\%$ \\
 & 995 & 31.96 & 20.29 & 32.32 & 20.01 & $-0.28\%$ & $-0.41\%$ \\
 & 990 & 24.67 & 30.09 & 25.27 & 29.78 & $+0.21\%$ & $+0.36\%$ \\
 & 985 & 20.40 & 36.69 & 20.91 & 36.28 & $+1.04\%$ & $+0.35\%$ \\
 & 980 & 18.03 & 42.49 & 18.24 & 42.43 & $+0.03\%$ & $+0.13\%$ \\
 & 975 & 16.03 & 46.39 & 16.39 & 46.53 & $+1.04\%$ & $+0.45\%$ \\
 & 970 & 14.58 & 50.14 & 14.76 & 49.69 & $+0.07\%$ & $+0.05\%$ \\
\midrule
BAF Var-III & 999 & 36.59 & 4.52 & 38.27 & 4.68 & $+0.04\%$ & $+1.59\%$ \\
 & 995 & 21.17 & 15.12 & 21.60 & 15.05 & $0.00\%$ & $+0.01\%$ \\
 & 990 & 16.54 & 23.87 & 16.96 & 23.57 & $-0.72\%$ & $+0.02\%$ \\
 & 985 & 13.95 & 30.13 & 14.32 & 29.83 & $+1.09\%$ & $+0.26\%$ \\
 & 980 & 12.45 & 35.82 & 12.53 & 35.52 & $-0.15\%$ & $-0.08\%$ \\
 & 975 & 11.10 & 39.63 & 11.30 & 39.59 & $+0.08\%$ & $+0.19\%$ \\
 & 970 & 10.11 & 43.06 & 10.22 & 42.61 & $-0.08\%$ & $-0.01\%$ \\
\midrule
BAF Var-IV & 999 & 36.59 & 4.52 & 38.27 & 4.68 & $+0.04\%$ & $+1.59\%$ \\
 & 995 & 21.17 & 15.12 & 21.60 & 15.05 & $0.00\%$ & $+0.01\%$ \\
 & 990 & 16.54 & 23.87 & 16.96 & 23.57 & $-0.72\%$ & $+0.02\%$ \\
 & 985 & 13.95 & 30.13 & 14.32 & 29.83 & $+1.09\%$ & $+0.26\%$ \\
 & 980 & 12.45 & 35.82 & 12.53 & 35.52 & $-0.15\%$ & $-0.08\%$ \\
 & 975 & 11.10 & 39.63 & 11.30 & 39.59 & $+0.08\%$ & $+0.19\%$ \\
 & 970 & 10.11 & 43.06 & 10.22 & 42.61 & $-0.08\%$ & $-0.01\%$ \\
\midrule
BAF Var-V & 999 & 40.36 & 5.05 & 40.54 & 5.09 & $+0.04\%$ & $+0.18\%$ \\
 & 995 & 22.43 & 16.48 & 22.76 & 16.29 & $-0.04\%$ & $-0.07\%$ \\
 & 990 & 17.52 & 25.45 & 17.76 & 25.00 & $+0.41\%$ & $+0.11\%$ \\
 & 985 & 14.74 & 31.86 & 15.03 & 31.18 & $+0.04\%$ & $+0.01\%$ \\
 & 980 & 13.16 & 37.78 & 13.13 & 37.18 & $-1.09\%$ & $-0.07\%$ \\
 & 975 & 11.82 & 42.16 & 11.98 & 41.67 & $+0.34\%$ & $+0.15\%$ \\
 & 970 & 10.79 & 45.59 & 10.89 & 45.06 & $-0.64\%$ & $-0.17\%$ \\
\bottomrule
\end{tabular}
\end{table*}

\section{D-Industrial per OOT Month}
\label{sec:app_month}

Table~\ref{tab:month} breaks the D-Industrial results of the all-3 ensemble down by OOT month, with
the cut-offs and every table fixed on the adaptation window and precision, recall, and the checks
computed within each month. March covers only its first seven days, so its per-cut-off values rest on
few frauds.

\begin{table*}[t]
\centering
\footnotesize
\caption{D-Industrial per OOT month, all-3 ensemble: $\Delta$R@P (top) and $\Delta$P@R (bottom) at each
  evaluation cut-off $t_b$ and their average (\%).}
\label{tab:month}
\begin{tabular}{lrrrrrrr}
\toprule
Month & 998 & 997 & 996 & 995 & 994 & 993 & Avg \\
\midrule
\multicolumn{8}{@{}l}{\textit{$\Delta$R@P}} \\
Jan 2026 & $+6.10\%$ & $+1.72\%$ & $+0.47\%$ & $+2.27\%$ & $+1.56\%$ & $+2.27\%$ & $+2.40\%$ \\
Feb 2026 & $+2.29\%$ & $+5.01\%$ & $+3.48\%$ & $+3.16\%$ & $+0.98\%$ & $+1.74\%$ & $+2.77\%$ \\
Mar 2026 & $-0.79\%$ & $+4.74\%$ & $+3.56\%$ & $+1.98\%$ & $+0.79\%$ & $-0.40\%$ & $+1.65\%$ \\
\midrule
\multicolumn{8}{@{}l}{\textit{$\Delta$P@R}} \\
Jan 2026 & $+2.72\%$ & $+2.37\%$ & $+0.88\%$ & $+1.64\%$ & $+2.35\%$ & $+1.10\%$ & $+1.84\%$ \\
Feb 2026 & $+1.13\%$ & $+2.96\%$ & $+2.78\%$ & $+2.08\%$ & $+0.84\%$ & $+0.84\%$ & $+1.77\%$ \\
Mar 2026 & $-0.28\%$ & $+2.00\%$ & $+1.91\%$ & $+1.64\%$ & $+1.78\%$ & $-0.61\%$ & $+1.07\%$ \\
\bottomrule
\end{tabular}
\end{table*}

\end{document}